\documentclass[a4paper, 10pt, conference]{ieeeconf}      
\usepackage{FG2018}

\FGfinalcopy 

\IEEEoverridecommandlockouts                              
\usepackage{algorithmicx}
\usepackage{algorithm}
\usepackage{algpseudocode}
\usepackage{amsmath}
\usepackage{amsfonts}
\usepackage{graphicx}
\usepackage{import}
\usepackage{mathtools}
\usepackage{interval}

\newcommand{\argmax}{\operatornamewithlimits{arg\,max}}

\providecommand{\norm}[1]{\lVert#1\rVert}

\def\R{\mathbb{R}}
\def\N{\mathcal{N}}

\def\GP{\mathcal{GP}}

\def\FGPaperID{XX} 

\title{\LARGE \bf
Morphable Face Models - An Open Framework
}

\author{\parbox{16cm}{\centering
    {\large Thomas Gerig, Andreas Morel-Forster, Clemens Blumer, Bernhard Egger, Marcel L\"uthi, Sandro~Sch\"onborn and~Thomas Vetter}\\
    {\normalsize
    Gravis Research Group, Department for Mathematics and Computer Science, University of Basel\\}}
    \thanks{This work was not supported by any organization.}
}

\begin{document}

\ifFGfinal
\thispagestyle{empty}
\pagestyle{empty}
\else
\author{Anonymous FG 2017 submission\\-- DO NOT DISTRIBUTE --\\}
\pagestyle{plain}
\fi
\maketitle

\begin{abstract}

In this paper, we present a novel open-source pipeline for face registration based on Gaussian processes as well as an application to face image analysis. Non-rigid registration of faces is significant for many applications in computer vision, such as the construction of 3D Morphable face models (3DMMs). Gaussian Process Morphable Models (GPMMs) unify a variety of non-rigid deformation models with B-splines and PCA models as examples. GPMM separate problem specific requirements from the registration algorithm by incorporating domain-specific adaptions as a prior model. The novelties of this paper are the following: (i) We present a strategy and modeling technique for face registration that considers symmetry, multi-scale and spatially-varying details. The registration is applied to neutral faces and facial expressions. (ii) We release an open-source software framework for registration and model-building, demonstrated on the publicly available BU3D-FE database. The released pipeline also contains an implementation of an Analysis-by-Synthesis model adaption of 2D face images, tested on the Multi-PIE and LFW database. This enables the community to reproduce, evaluate and compare the individual steps of registration to model-building and 3D/2D model fitting. (iii) Along with the framework release, we publish a new version of the Basel Face Model (BFM-2017) with an improved age distribution and an additional facial expression model.

\end{abstract}

\section{Introduction}\label{sec:introduction}

A popular approach for modeling the variability of human faces is the Morphable Model. Besides its capability to analyze a population of shapes, its primary purpose is the reconstruction of the 3D face surface from single face images as proposed in \cite{blanz1999morphable}. Crucial for the construction of the morphable model is a dense correspondence between the points of the training surfaces. This is established with shape registration, which deforms a reference shape to match a given target shape. The quality of the provided target shapes heavily depends on the scanning process itself and is often corrupted with artifacts (hair, eyebrows), missing data and outliers. Facial expressions add another layer of complexity, which is mainly associated with the mouth opening and closing. Also, typical faces contain variability on different levels of detail and symmetrically correlating features. Algorithms, tailored for faces, such as \cite{amberg2007optimal}, successfully deal with these domain specific issues. One fundamental problem, however, is that the prior assumptions about the data are not separated from the registration algorithm itself. This results in a complex mix of concepts, all implemented as algorithmic components in the registration algorithm. Recently, \cite{luethi2016gpmm} proposed a framework, based on Gaussian processes, which enables to model prior assumptions about the registration problem decoupled from the registration algorithm itself. The GPMM framework models deformations from a reference surface to a target surface as a Gaussian process $\GP(\mu, k)$ with mean function $\boldsymbol\mu\colon \Omega \to \R^3$ and covariance (or kernel) function $k \colon \Omega \times \Omega \to \R^{3 \times 3}$. The kernel function $k$ describes the type of deformations, and the Gaussian process itself models a probability distribution over the deformations, which is also called prior model. 

In this paper, we derive a method for face registration based on Gaussian Process Morphable Models, where face specific domain knowledge is modeled with a Gaussian process. This approach has the following advantages:

\begin{itemize}
    \item The method is conceptually simple because problem specific adaptions are decoupled from the registration algorithm.
    \item Domain knowledge is modeled intuitively using building blocks in terms of kernels.
    \item Extending the model does not require changing the registration algorithm.
    \item As the deformation prior is generative, random samples can be drawn to check modeling assumptions visually.
\end{itemize}

We show how to build a prior model for faces incrementally: 

\textbf{1)} The geometric variability of the face can be decoupled into multiple levels of detail. We model this variability with multi-scale B-spline kernels and propose an adaption scheme to damp the predefined regions on different deformation scales spatially.

\textbf{2)} Facial shapes are nearly mirror symmetric. We model this by modeling symmetry with a mirror symmetric kernel.

\textbf{3)} We propose a simple statistical shape model kernel built from facial expression prototypes to model the opening and closing of the mouth.

\textbf{4)} To build a shape and texture model from the registered data, we propose a model-building method, which also handles regions with missing data.

A further primary purpose of this work is full reproducibility on publicly available data. We release the full face registration and model-building pipeline together with experiments on the model adaption of a single 2D image. By releasing the complete pipeline, tested on publicly available data, we provide full reproducibility for all the individual steps and the end-result of the pipeline. 

We also release a new Basel Face Model (BFM-2017). The model contains facial expressions, is based on an improved age distribution compared to the model published by \cite{paysan20093d} and is built with training samples that have been recorded in a well-controlled environment.

The paper is organized as follows: Section~\ref{sec:related_work} describes work related to this topic. Gaussian processes and their usage for modeling deformation priors are described in Section~\ref{sec:gpmm} and \ref{sec:combiningkernels}. In Section~\ref{sec:modelling} we propose a new kernel for face registration. The registration pipeline itself is explained in Section~\ref{sec:registration}. Section~\ref{sec:data} explains the different datasets that are used for this work. In Section~\ref{sec:results} the quantitative and qualitative results of the BU3D-FE registration and a model adaption application of single 2D images are shown. At last, our conclusions are drawn in Section~\ref{sec:conclusion}.

\section{Related Work}\label{sec:related_work}

The iterative closest point algorithm (ICP) \cite{besl1992method} and its non-rigid extension (NICP), introduced by \cite{amberg2007optimal} and \cite{allen2003space} are the most popular algorithms used for establishing the correspondence of 3D face shapes (\cite{paysan20093d},\cite{booth2017large},\cite{alyuz2008},\cite{Cheng20173},\cite{hasler2009}). Extending the non-rigid ICP to a specific problem domain or data-set requires changes in point search heuristics and stiffness weights, which makes the method complicated to adapt in practice. Additional extensions the NICP algorithm have also been proposed: \cite{alyuz2008} introduced independent local components for the NICP algorithm to handle the difficulty of facial expressions. In \cite{Cheng20173}, local statistical models, trained from registered data, are embedded as constraints in the NICP algorithm. In addition to NICP, alternative approaches for face registration have been proposed: In \cite{xiaolei2006}, a registration algorithm with a B-spline based deformation model is shown. In \cite{tena2006}, the authors propose an algorithm based on thin-plate-splines, which handles different levels of detail and mirror symmetry. \cite{bronstein2007expression} propose to model facial expressions with mouth opening as isometric deformations on the face surface. \cite{Salazar2014} handle the expression problem by fitting an expression model of blendshapes before the shape registration step. In case of model-building, the BFM~\cite{paysan20093d} is the most used Morphable Model in literature, and it was built on 200 neutral faces using NICP. Recently a large scale Morphable Model built from 10'000 faces has been proposed using NICP for registration \cite{booth2017large}). Both those models lack facial expressions. The Surrey face model contains facial expressions, which are built from 6 blendshapes and provides multiple resolutions of their shape model \cite{huber2016multiresolution}. A statistical shape model (no color) was built on the BU-3DFE face database using a multilinear expression model \cite{bolkart2016robust}.
After registration and model-building, we demonstrate the applicability of the model with an inverse rendering application of 2D face images. Unlike the approach by \cite{Schoenborn2017}, which is used in this work, most methods only recover shape but ignore color and illumination. An overview over current inverse rendering techniques is contained in \cite{Schoenborn2017}. A recent publication presents an end-to-end learning of rendering and model adaptation incorporating a 3DMM \cite{mofa2017}.

\section{Method}\label{sec:method}

\subsection{Gaussian Processes for Face Registration}\label{sec:gpmm}
For establishing correspondence among the individual surfaces, we use an approach for non-rigid registration proposed by \cite{luethi2016gpmm}. In this approach, registration is formulated as a model-fitting problem, where the model is obtained by modelling the possible deformation of a reference surface, using a Gaussian process. More precisely, let $\Gamma_R \subset \R^3$ be the reference surface, which should be in our case be a face mesh of high quality and anatomically normal shape. To define the model, we assume that any target face $\Gamma_T \subset \R^3$ can be written as a deformed version of this reference shape with a deformation field $u : \Gamma_R \to \R^d$, i.e.

\begin{equation}
\Gamma_T =  \{x + u(x) | x \in \Gamma_R \}.
\end{equation}

We define a prior over the possible deformations using a Gaussian process $u \sim GP(\mu, k)$, where $\mu : \Gamma_R \to \R^3$ is a mean function and $k : \Gamma_R \times \Gamma_R \to \R^{3 \times 3}$ is a covariance function. The mean function defines the average deformation from the reference that we expect (which is typically the zero function, assuming that the reference is an average face) and the covariance function defines the characteristics of the allowed deformations. The resulting model is a fully probabilistic model over face shapes. To see this, notice that for every face $\Gamma_T$ we can now assign a probability $p(\Gamma_T) = p(u) = GP(\mu, k)$ determined by the Gaussian process. Conceptually, the registration problem is now cast as the MAP problem

\begin{equation}
\argmax_u p(u |\Gamma_T,  \Gamma_R)  = \argmax_u p(u) p(\Gamma_T |u , \Gamma_R).
\end{equation}

To turn this conceptual problem into a practical one, we need to fix the likelihood function $p(\Gamma_T | u, \Gamma_R)$
and find a strategy to optimize the problem.
For the likelihood function we define the distance between a point $x$ and the target surface as

\begin{equation}\label{eq:distance}
d_{\Gamma_T, \Gamma_R}(x_i,u) = \rho(CP_{\Gamma_T}(x_i + u(x_i)) - (x_i + u(x_i)))).
\end{equation}

with $\rho$ as a loss function and $CP_{\Gamma_T}(x)$ as the closest point on surface $\Gamma_T$ to x:

\begin{equation}\label{eq:closest-point}
CP_{\Gamma_T}(x) = \min_{x_t \in \Gamma_T} \norm{x - x_t}^2.
\end{equation}

Assuming independence of the errors at every vertex, we obtain the likelihood function:
\begin{equation}
p(\Gamma_T|u, \Gamma_R) = \frac{1}{Z}\prod_{x_i \in \Gamma_R} exp(-\frac{d_{\Gamma_T, \Gamma_R}(x_i,u))^2}{\sigma^2})
\end{equation}
To parameterize the infinite dimensional optimization problem, \cite{luethi2016gpmm} propose to approximate the model using  a truncated Karhunen-Lo\`eve expansion. This leads to a parametric model $\tilde{u}(\alpha, x)$ of the form
\begin{equation}\label{eq:lowrank}
\tilde{u}(\alpha,x) :=  \mu(x) + \sum_{i=1}^r \alpha_i \sqrt{\lambda}_i \phi_i(x), \, \alpha_i \sim \N(0,1),
\end{equation}
where $\lambda_i \in \R$ are weights and $\phi_i : \Gamma_R \to \R^3$ corresponding basis function.
Note that under this approximation, the probability of observing the target shape $\Gamma_T$ is completely determined by the parameter vector $\alpha=(\alpha_1, \ldots, \alpha_r)^T$ and thus

\begin{align}
\begin{split}
& p(\Gamma_T) = p(\tilde{u}) = p(\alpha) \\ 
& = N(0, I_{r \times r}) = \frac{1}{Z}\exp(-\norm{\alpha}^2).
\end{split}
\end{align}

The final registration problem is then:

\begin{equation}\label{eq:reg-map}
\begin{multlined}
\argmax_{\alpha} p(\alpha) p(\Gamma_T | \alpha , \Gamma_R) = \\
\frac{1}{Z} \exp(-\norm{\alpha}^2)  \\ 
\prod_{x_i \in \Gamma_R} \exp(-\frac{d_{\Gamma_T, \Gamma_R}(x_i,\tilde{u}(\alpha, x_i)))^2}{\sigma^2})
\end{multlined}
\end{equation}

This is a parametric optimization problem, which can be approached using standard optimization algorithms. For this work, an implementation of LBGFS~\cite{nocedal1980updating} was used. 

\subsection{Combining Kernels}\label{sec:combiningkernels}

The covariance function $k : \Gamma_R \times \Gamma_R \to \R^{3\times 3}$, which
is also referred to as the kernel function, defines the characteristics of the deformations.
Let $g, h :\Omega \times \Omega \to \R$ be two symmetric positive semi-definite kernels and $f : \Omega \to \R$ an arbitrary function. Then the following rules can be used to generate new positive semi-definite kernels, which is well described in \cite{duvenaud-thesis-2014}:

\begin{align}\label{eq:combine}
\begin{split}
k(x,x') &= g(x,x′)+h(x,x′) \\
k(x,x') &= \alpha g(x,x′), \alpha \in \R_+ \\
k(x,x') &= g(x,x')h(x,x') \\
k(x,x') &= f(x)f(x').
\end{split}
\end{align}

\subsection{A Shape Prior tailored for Face Registration}\label{sec:modelling}

In this section we show how to build a deformation prior for face registration. As the reference surface $\Gamma_R$ we have chosen the mean shape of the Basel Face Model.
It is therefore a good assumption to choose the mean deformation to be the zero function,
\begin{equation}
\mu(x) = (0,0,0)^T, \,  x \in \Gamma_R.
\end{equation}

\subsubsection{Multi-scale Deformations}

\begin{figure*}
    \begin{center}
        \includegraphics[width=0.7\linewidth]{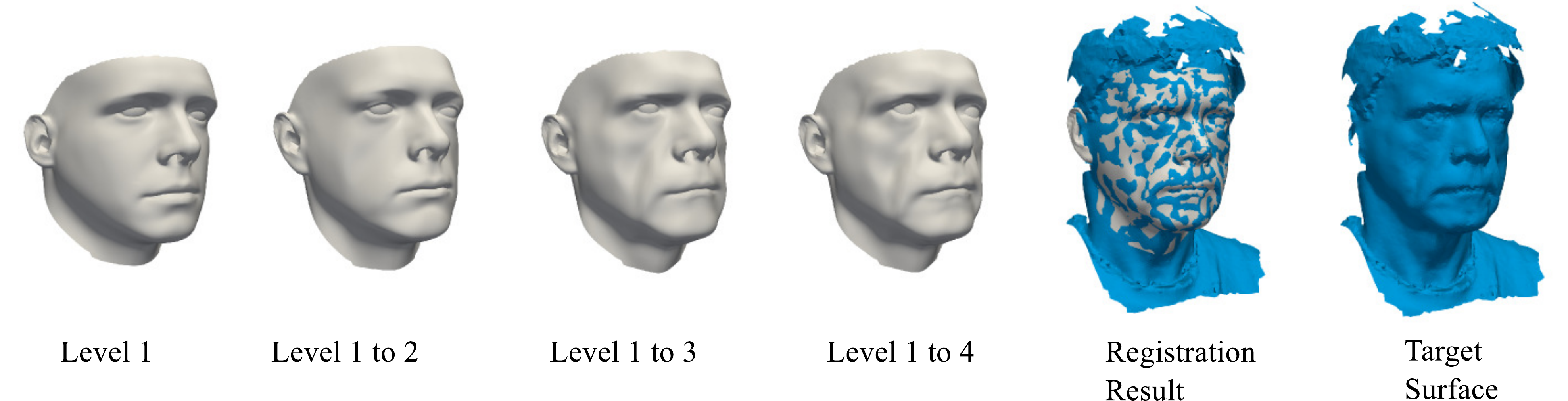}
    \end{center}
    \caption{Multi-scale registration example. \emph{On the left}, the registration results with different levels of details are shown. Shape~\textbf{Level 1} is the result of the registration simply with the lowest scale. From \textbf{Level 1 to Level 4} the number of levels is increased, which leads to details on finer scales. It is to point out that the changes in \textbf{Level 1 to Level 2} are of coarse nature and represent head shape and coarse positioning of nose and eyes. In \textbf{Level 3 to Level 4}, fine deformation changes, such as folds, eye and nose shape are visible. \emph{On the right side}, the full registration result (all scales) in comparison to the target shape in blue is visualized.}
    \label{figMultiscale}
\end{figure*}

As the basis of the model, we chose the multi-scale B-spline kernel, introduced in \cite{opfer2006multiscale}. Given a univariate third order B-spline $b_3$ and the function $\psi(x) = b_3(x_1)b_3(x_2)b_3(x_3)$, the kernel reads

\begin{equation}\label{eq:bspsingle}
k_j(x,x') = \sum_{k \in \mathbb{Z}^d}2^{2-j}\psi\big(2^{j}x-k\big)\psi\big(2^{j}x'-k\big)
\end{equation}
with $k$ evaluated in the support of the B-spline. The multiple scales are defined as
\begin{equation}
k_{\text{BSp}}(x,x') = I_{3x3} \sum_{j=\underline{j}}^{\overline{j}} s_j k_j(x,x')
\end{equation}
with level $j$ from coarse $\underline{j}$ to fine $\overline{j}$ and multiplied by the identity matrix $I$ to get a matrix valued kernel. The value $s$ is the deformation scale per level and thus defines how far the correlating points can deform. It is chosen such that coarse scale levels are able to deform more than finer levels. This kernel defines smoothly varying function on multiple scale-levels. The individual scales can be decoupled as a superposition of different levels as shown Figure~\ref{figMultiscale}. All scale layers combined to the fully detailed registration result are shown next to the target shape in blue. The increasing level of scale from Level~1 until Level~4 is shown for comparison. Level~1 is defined as $k_{BSp}(x,x')$ with $j=1~to~1$ and Level~4 as $j=1~to~4$. While in Level~1 and Level~2 coarse details are globally adapted, skin details and eye shape are deformed with small scale deformations (Level~4). 

\subsubsection{Spatially Varying Scales}

\begin{figure}
    \begin{center}
        \includegraphics[width=0.7\linewidth]{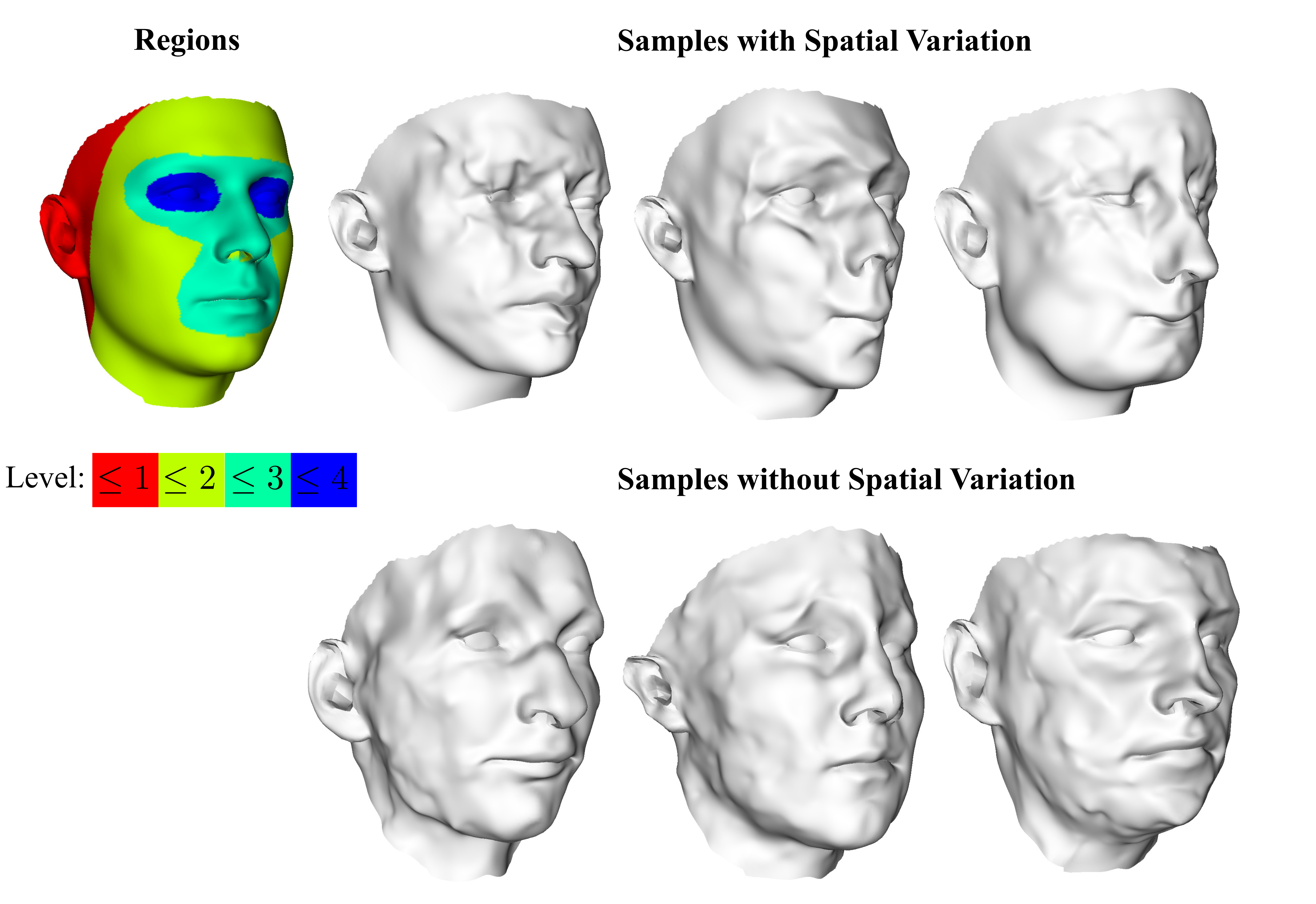}
    \end{center}
    \caption{Comparison of a spatially-varying multi-scale kernel (top) and a kernel without spatially-varying scales (bottom). The region map indicates where the scale levels are active. The red area restricts the kernel to the lowest deformation level, which results in coarse scale deformations around the ears. The yellow region around the cheeks allows more levels than red, which yields more details. From green to blue the amount of levels and the details increase, which is especially visible around the eyes. For comparison we show samples of a kernel without spatially varying, where all deformation scales are present over the whole domain.}
    \label{fig:spatially}
\end{figure}

Typical face shapes contain small scale variability around the eyes and mouth, but are rather smooth around the cheeks. Therefore we have divided the face into smooth regions and combine this information with the multi-scale B-spline kernel. This leads to a model with small scale deformations around the eyes and mouth region, while the cheeks are still restricted to smooth, large scale deformations (see Figure~\ref{fig:spatially}: Regions). 

\begin{equation}\label{eq:multi-scale-kernel-sv}
k_{\text{svms}}(x,x') = \sum_{j=\underline{j}}^{\overline{j}} \chi^j(x) \chi^j(x') k_j(x,x')
\end{equation}

where $\chi^j : \Gamma_R \to \interval{0}{1}$ are smooth indicator functions that determine if the kernel is active (i.e. $\chi^j(x) = 1$) at location $x$ for level $j$ and $k_j(x,x')$ is a single scale B-spline kernel defined as in (\ref{eq:bspsingle}). In Figure~\ref{fig:spatially}, random samples of the described kernel are shown in comparison to samples from the standard non-varying kernel.

\subsubsection{Symmetry}

A face is nearly mirror symmetric, which should be reflected in the model. We follow an approach to define axial symmetric kernels proposed in \cite{morelforster2017generative}. Given an arbitrary scalar-valued kernel function $k: \Gamma_R \times \Gamma_R \to \R$, the authors have shown how to define valid matrix-valued kernel $k: \Gamma_R \times \Gamma_R \to \R^3$ for modelling mirror-symmetric deformation fields. The symmetric covariance function is given by

\begin{equation}
k_{sym}(x,x';k) = I k(x,x') + \bar{I} k(x,\bar{x}')\text{ ,\ \ \ }
\end{equation}
where $I$ is the $3\times3$ identity matrix and
\begin{equation}
\bar{I} = \begin{Bmatrix}
-1 & 0 & 0 \\
0 & 1 & 0 \\
0 & 0 & 1
\end{Bmatrix} \ , \ \ \bar{x}' = \begin{Bmatrix}
-x'_1\\
x'_2\\
x'_3
\end{Bmatrix} \ .
\end{equation}

\begin{figure}
    \begin{center}
        \includegraphics[width=0.7\linewidth]{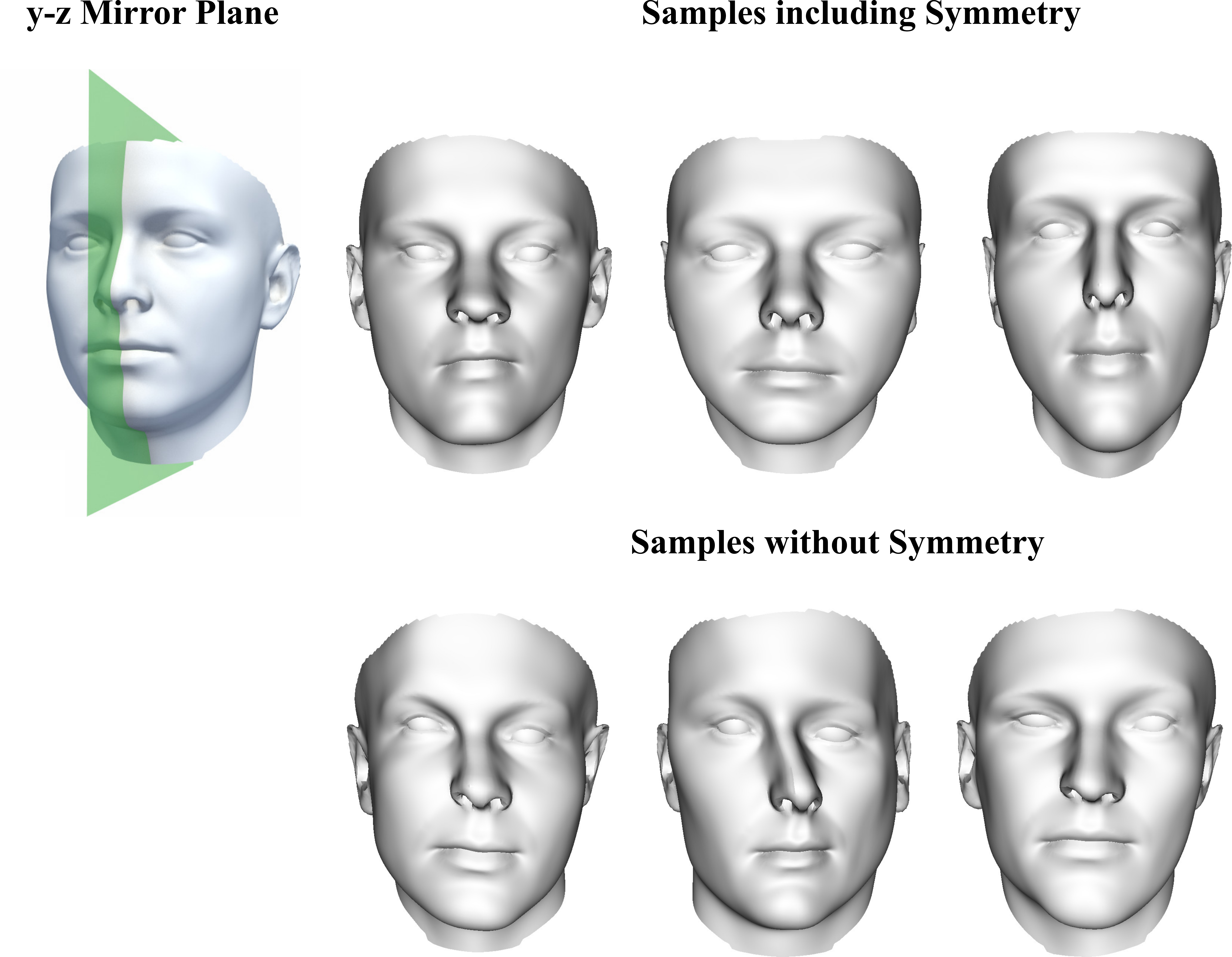}
    \end{center}
    \caption{A comparison between a normal and a symmetrized kernel. The results of the symmetry kernel are illustrated on the top row. The samples illustrated in the bottom row do not represent realistic face examples because of strong asymmetry.}
    \label{fig:symmetry}
\end{figure}

Intuitively, this construction takes a definition $k(x,x)$ of how the function values at two points $x,x'$ on the surface $\Gamma_R$ are correlated. Then the correlations of the three components of the resulting deformation field are constructed by multiplying with the identity matrix $I$. To achieve mirror-symmetry, the minus sign is introduced in the first component to ensure that correlation between two points, which are on the opposite side of the symmetry plane lead to the inverse correlations. The symmetry is integrated in the face model by combining symmetric and asymmetric deformations to make the face samples look more realistic. In Figure~\ref{fig:symmetry} a comparison between a normal and a symmetrized kernel is visualized.

\subsubsection{Core Expression Model}

\begin{figure}
    \begin{center}
        \includegraphics[width=0.9\linewidth]{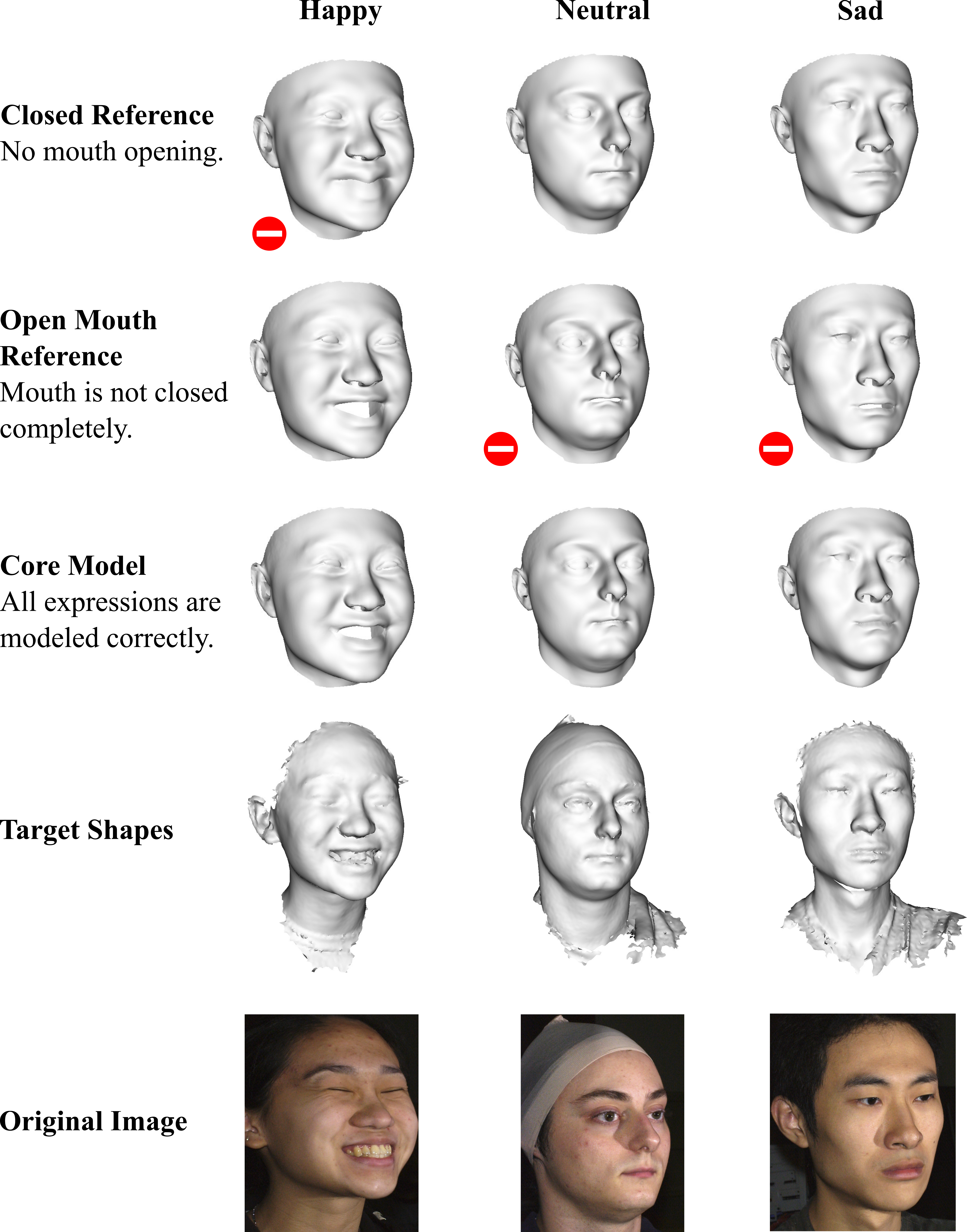}
    \end{center}
    \caption{Closed and open-mouth registration examples with and without a core expression model. The top row shows registrations to three expression examples using a reference shape with a closed mouth. The middle row shows that a reference shape with an open mouth leads to bad results with the neutral and sad example. The registration using the core model yields successful results for all three examples.}
    \label{fig:coremodel}
\end{figure}

In facial expressions, the opening and closing of the mouth cannot be modelled simply with a smooth kernel, such as a B-spline or radial basis function. Since the points on the upper and lower lip are close, they correlate strongly, which hinders an opening deformation. One approach is the usage of a new reference with an open mouth. However, the registration with multiple templates is inconvenient in practice. The second row in Figure~\ref{fig:coremodel} visualizes the registration using an open mouth reference. Although it gives perfect results for open mouth registrations, the mouth does not close properly for neutral faces. To build a model that can cope with both situations, we combine a simple statistical shape model with the previously described prior model. To build this facial expression kernel we make use of the facial expression reference shapes (anger, disgust, fear, happy, sad, surprise) $u_i$ to compute the mean

\begin{equation}
\mu_{sm}(x) = \frac{1}{n} \sum_{i=1}^n u_i(x)
\end{equation}
and covariance function
\begin{align} \label{eq:smkernel}
\begin{split}
 &k_{sm}(x, x') = \frac{1}{n-1} \\
 &\sum_{i=1}^n (u_i(x) - \mu_{sm}(x)) (u_i(x') - \mu_{sm}(x'))^T.
 \end{split}
\end{align}

The facial expression kernel $k_{ssm}(x,x')$ can be combined with another kernel according to the rules in (\ref{eq:combine}), which again results in a valid kernel function. For the face registration we use the kernel formulated in the previous sections $k_{svms}(x,x')$ in combination with the core model. The final deformation model is defined by symmetrizing the spatially-varying kernel
\begin{equation}
k_{sym}(x,x') = k(x,x';k_{svms}(x,x'))
\end{equation}
and augment the function with the facial expression kernel:
\begin{equation}
k_{expr}(x,x') = k_{sm}(x,x') + k_{sym}(x,x')
\end{equation}

In Figure~\ref{fig:coremodel}, bottom row, the results of a registration using $k_{expr}(x,x')$ are shown. All the test-cases, the closed, as well as the open mouth samples, have been accurately registered.

\subsection{The registration algorithm}\label{sec:registration}

\begin{algorithm}
\begin{algorithmic}[1]
\Procedure{Registration}{}
\State  Compute posterior model $GP(\mu_p, k_p)$ for landmarks
\State $\alpha \gets 0^n$ \Comment(initial solution)
\For{$\eta \gets (1e-1, 1e-2, \ldots, 1e-5)$}
\State $\Gamma(\alpha) \gets \text{Current best fit (surface)}$
\State Find and discard outliers using $\Gamma(\alpha)$
\If{line annotations available}
\State Find matching line landmarks using $\Gamma(\alpha)$
\State Compute posterior model for lines
\EndIf
\State $\alpha \gets $solution to \eqref{eq:reg-map} with regularization weight $\eta$
\EndFor
\EndProcedure
\end{algorithmic}
\caption{High level overview: Registration procedure}\label{alg:reg-algo}
\end{algorithm}

So far we have described how the registration algorithm works in principle: We formulate a Gaussian process model $GP(\mu, k)$ as a prior and minimize (\ref{eq:reg-map}). The steps are summarized in Algorithm~\ref{alg:reg-algo}. In the first step we make use of the provided landmark points in the registration. Gaussian process morphable models \cite{luethi2016gpmm} make it possible to include those landmarks directly into the prior by considering the deformation $\hat{u}^i := l_T^i - l_R^i$ between a landmark pair $l_R^i, l_T^i$, as a noisy observation of the true deformation $\hat{u}$, i.e.\
\begin{equation} \label{eq:lms}
u(l_R^i) = \hat{u}^i + \epsilon, \, \epsilon \sim N(0, \sigma I_{3 \times 3}),
\end{equation}
and applying Gaussian regression to it, as described in \cite{luethi2016gpmm}. The resulting posterior distribution
assigns a low probability to any deformation $u$ that does not match the specified landmarks (up to the specified uncertainty $\sigma$). The posterior model is again a Gaussian process, and thus can be used instead of the original prior, without changing the algorithm. The registration problem \eqref{eq:reg-map} is optimized in different steps with decreasing regularization weights. In each step, all the points of the model for which the current fit is further away from the target surface than some predefined threshold or whose closest point is a boundary point (indicating a hole in the target surface) are eliminated from the optimization.

To describe the distance metric that has been used, we denote $CP_{\Gamma_T}(x)$ as defined in (\ref{eq:closest-point}) and calculate the distance (\ref{eq:distance}) with $\rho$ as the Huber loss function defined by
\begin{equation}
\rho(x) = \left \{ \begin{array}{cc} \frac{x^2}{2} & \text{if} |x| < k \\ k(|x| - k/2) & \text{otherwise.} \end{array} \right .
\end{equation}

\subsection{Building the Morphable Model}

\subsubsection{Missing Data}

To build a color model, the closest corresponding color value of the target mesh is extracted at all points on of the registered mesh. Since the target scans are often incomplete and contain holes, not every point in the registration can be assigned a color value. To address this issue we introduce a binary indicator variable $z\in \{0,1\}$ to specify whether a reliable color at a point $x$ exists or not. We then compute the color mean using only the available colors

\begin{equation}
\mu_{md}(x) = \frac{1}{\sum_{i=1}^n z_i} \sum_{i=1}^n z_i u_i(x) .
\end{equation}

When estimating the covariance function for the color model we use an additional kernel $k_{cs}(x,x')$ to express our prior similar to the smoothness assumption for the shape surface.\footnote{In practice, we use a single level square exponential kernel with a scaling of $1.0e^{-4}$ and a correlation of $\sigma=10$, where the units are millimeters.} Based on $z_i$ and $z'_i$ we use either the empirical covariance or the the covariance specified by the prior kernel. The full color covariance function handling missing data is then
\begin{align}\label{eq:smkernel}
\begin{split}
 &k_{md}(x, x') = \\
 &\frac{1}{n-1} \sum_{i=1}^n ( z_i z'_i c(x,x') + (1- z_i z'_i) k_{cs}(x,x') )
\end{split}
\end{align}

with $c(x,x')$ as the same term used in the empirical estimate:

\begin{equation}
c(x,x') =  (u_i(x) - \mu_{md}(x)) (u_i(x') - \mu_{md}(x'))^T.
\end{equation}

\subsubsection{Expression Model}

We extend the original face model to a multi-linear model to handle expressions as described in \cite{amberg2008expression}. The multi-linear statistical model consists of two independent models for face shape and face color as well as an additional model for the deformations by facial expression. Facial expression is modeled as a difference from the neutral face shape.

\section{Data}\label{sec:data}

\subsection{BU-3DFE Database}

The Binghamton University 3D Facial Expression Database (BU-3DFE) \cite{yin20063d} has neutral and expression scans of 100 individuals. Per individual it contains 6 facial expressions with 4 levels of strength. For the registration pipeline we use the raw data without cropping and a single expression strength (Level 4). 

For the F3D data 83 detected landmarks are given. The RAW data has 5 landmarks. We used an ICP alignment to transfer the F3D landmarks onto the RAW data. Additionally, we clicked 23 landmarks for all neutral scans and expression scans of level 4 for the registration and used the F3D points for correspondence evaluation.

All scans have a texture file which is constructed from two pictures ($\pm 45$ degrees), see also the bottom row of Figure~\ref{fig:coremodel}. However, no ambient illumination was ensured, and many illumination effects (e.g., shadows on both sides of the nose or strong specular highlights) are visible. Additional disadvantages like make-up, facial hair or hair falling into the facial area do occur. We demonstrate the advantages of controlled data compared to the BU-3DFE data in Section~\ref{sec:results}.

\subsection{Basel Scans}

We used the face scans introduced in \cite{paysan20093d} which are scanned under a strictly controlled environment. For further details about the data we refer the reader to the original publication. In contrast to \cite{paysan20093d}, we used an improved age distribution which includes more people over 40 years. The advantages of using the Basel scans for building a high quality morphable face model are:

\begin{itemize}
 \item Strict setting for scanning: No make-up, beards or hair in the facial area.
 \item Number of scans: more individuals than in BU-3DFE.
 \item Texture quality: Ambient illumination and the texture in high resolution and good quality.
 \item Age: Known at scanning time.
 \item Expressions: 6 types (anger, disgust, fear, happy, sad, surprise), controlled conditions.
\end{itemize}

In case of the Basel scans, also contour lines are available. We have included them in the registration by calculating a posterior model to the closest points on the line before every registration step, as mentioned in Algorithm~\ref{alg:reg-algo}.

For the new Basel Face Model (BFM-2017), a more representative data distribution compared to the original BFM (see also \cite{paysan20093d}) has been selected. Also, a facial expression model has been included, built from 160 examples. The data has been chosen the following:

\begin{itemize}
    \item 100 male and 100 female shape examples.
    \item 100 male and 100 female color examples.
    \item 160 expression examples, equally distributed on expression types.
\end{itemize}

\begin{figure}
    \begin{center}
        \includegraphics[width=1.0\linewidth]{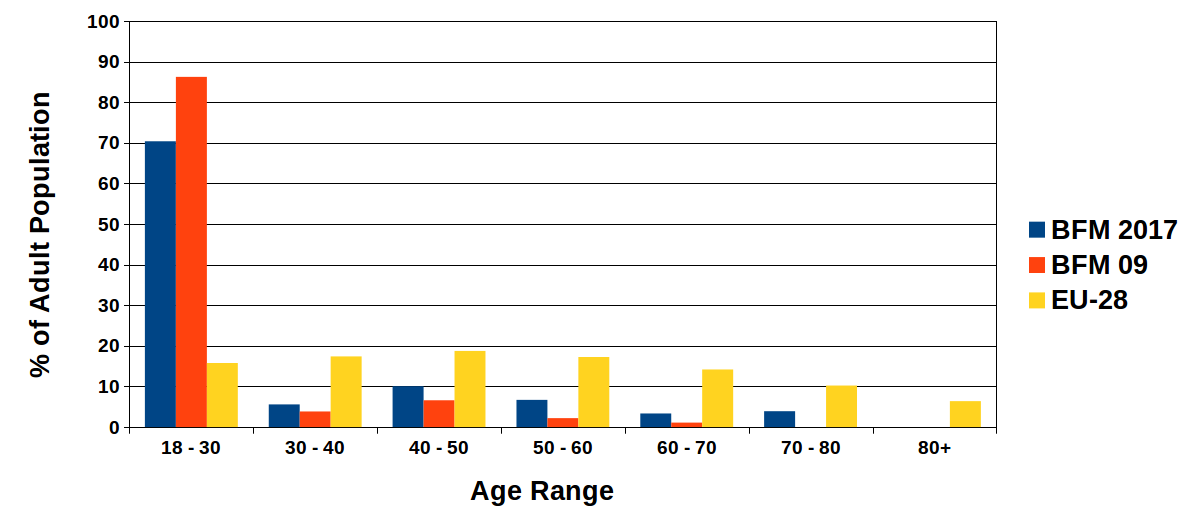}
    \end{center}
    \caption{Age distribution of the new model compared to the original BFM and to the European Union (EU-28 in 2013 \cite{EuroStat}). The lack of older people was corrected.}
    \label{fig:data:agedistribution}
\end{figure}

In Figure~\ref{fig:data:agedistribution} it is visible that the selected data is closer to the real age distribution (e.g., from the European Union). Moreover, the improved age distribution reflects the importance of people older than 40 years. This group has facial attributes like sacking and wrinkles, which young people (below 30) are mostly lacking off.

\section{Results}\label{sec:results}

\subsection{Landmark Evaluation}

\begin{table}

\begin{center}
    
\begin{tabular}{ |c|c|c| } 
 \hline
 Face Region & Dist [mm](Ours) & Dist [mm](Salazar)\\
 \hline
 Left Eyebrow & $4.69 \pm 4.64$ & $6.25\pm 1.84$ \\ 
 Right Eyebrow & $5.35 \pm 4.69$ & $6.75 \pm 3.51$\\
 Left Eye & $3.10 \pm 3.43$ &$ 3.25 \pm 1.84$ \\
 Right Eye & $3.33 \pm 3.53$ & $3.81 \pm 2.06$ \\
 Nose & $3.94 \pm 2.58$ & $3.96 \pm 2.22$\\
 Mouth & $3.66 \pm 3.13$ &$ 5.69 \pm 4.45$ \\
 Chin & $11.37 \pm 5.85$ &$ 7.22 \pm 4.73$\\
 Left Face & $12.52 \pm 6.04$ & $18.48 \pm 8.52 $ \\
 Right Face & $10.76 \pm 5.34$ & $17.36 \pm 9.17 $\\
 \hline
\end{tabular}
  \caption{The landmarks used for the evaluation correspond to the BU-3DFE database landmarks and are semantically sorted as in \cite{Salazar2014}. In this table, the average distance error between the estimated position of the registration result and the position provided by the BU-3DFE database is shown (the smaller, the better). The result is computed over all the registration results of all facial expressions. The results are compared to \cite{Salazar2014}.
}
\label{tab:landmarkdistance}
\end{center}

\end{table}

To provide a measure of the registration accuracy with the BU-3DFE database, we compare our registrations to the landmarks, which are provided with the BU-3DFE database. To evaluate an average distance error, the landmarks of the registrations are compared to the positions that are provided with the BU-3DFE dataset. In Table~\ref{tab:landmarkdistance}, the average distance error per region is shown. We sorted the BU-3DFE landmarks as described in \cite{Salazar2014}, to match their evaluation scheme. The proposed registration shows a similar correspondence as annotated in the BU-3DFE database and are on par with the evaluation in \cite{Salazar2014}. The high standard deviation is due to the fact that all expressions are evaluated together. Expressions, such as anger and fear heavily affect the shape of the eye and eyebrows, which in turn has impacts on the standard deviation. To enable a better comparison in future work, we provide our manually clicked landmarks on the reference mesh together with the source code.

\subsection{Inverse Rendering}\label{sec:fitting}
The original application of 3D Morphable Face Models proposed in \cite{blanz1999morphable} is an inverse rendering task. Inverse rendering aims to estimate all necessary parameters $\theta$ of an image formation process to generate a given target image. The full model consists of the statistical shape, color and expression model, a pinhole camera model as well as spherical harmonics for illumination modeling (\cite{basri2003lambertian,zivanov2013human}).
To complete the framework, we include an implementation of a recent technique to estimate the parameters from a single still image. The framework we are implementing is a fully probabilistic model adaptation framework \cite{Schoenborn2017} based on Markov chain Monte Carlo sampling. The face model is integrated as a prior on facial shape, expressions and color appearance into this model adaptation framework. Such a strong prior is necessary to be able to reconstruct the 3D shape from a single 2D image. We extended the model adaptation framework to handle facial expressions by including expression proposals like the ones for the shape and color coefficients as described in \cite{egger2017probabilistic}. This is the first publicly available implementation of a 3D Morphable Model adaptation framework in an Analysis-by-Synthesis setting including facial expressions.

\begin{figure*}
    \begin{center}
        \includegraphics[width=1\linewidth]{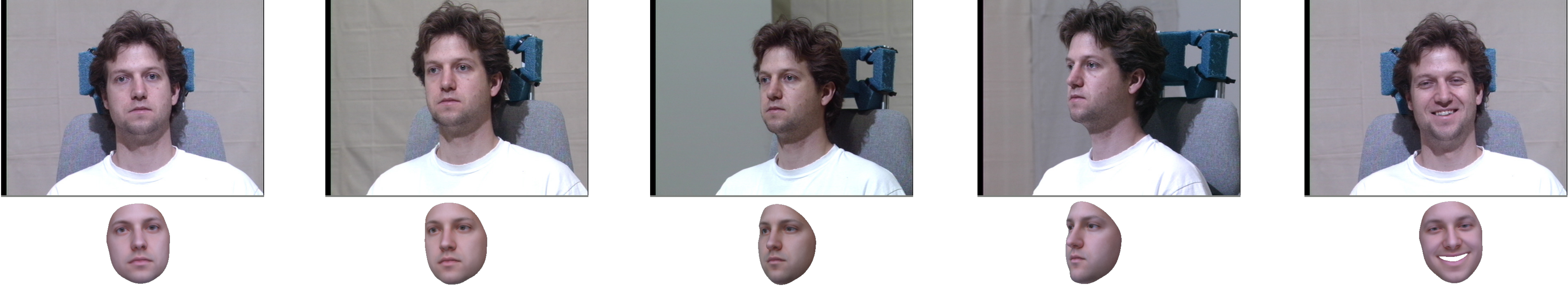}
    \end{center}
    \caption{Qualitative model adaption result on the Multi-PIE face database for the first of 249 individuals (compare Table~\ref{tab:multiPieRecognition}). The second row shows qualitative model adaptation results using the BFM-2017 model.}
    \label{fig:multiPieFits}
\end{figure*}

\begin{figure}
    \begin{center}
        \includegraphics[width=1\linewidth]{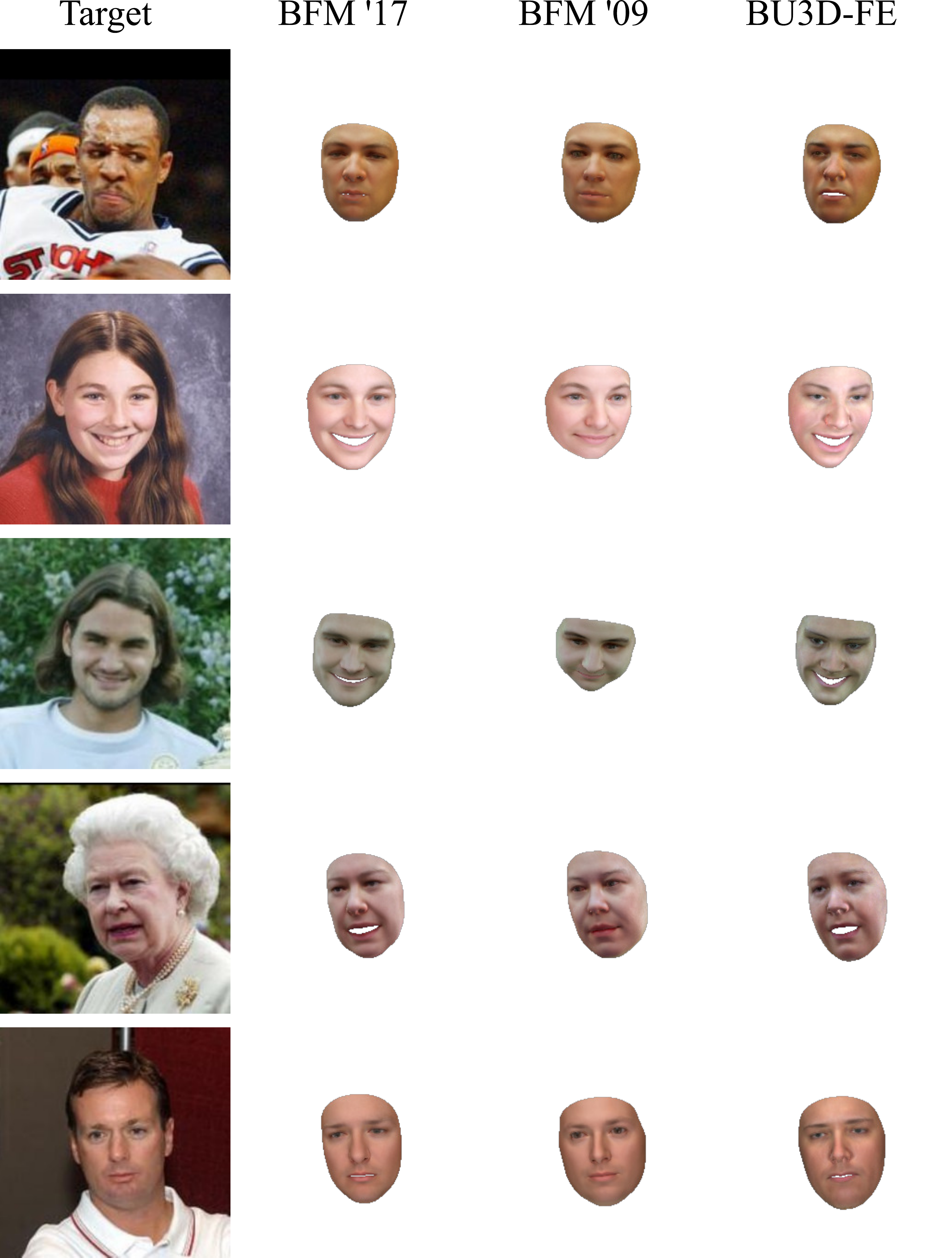}
    \end{center}
    \caption{Qualitative model adaptation results on the LFW database. On the left the original target image followed by the results obtained by the different models. The new model leads to consistently better results than the old one and the one built on lower quality data from the BU3D-FE dataset.}
    \label{fig:lfwFits}
\end{figure}

We present results of our face model adaptation method on the Multi-PIE database \cite{gross2010multi}. For our experiments we used the neutral and smiling photographs of 249 individuals in the first session in four poses (0$^{\circ}$ camera \texttt{051}, 15$^{\circ}$ camera \texttt{140}, 30$^{\circ}$ camera \texttt{130}, 45$^{\circ}$ camera \texttt{080}) under frontal illumination (illumination 16). We show the different poses and expressions together with there fitting results in Figure~\ref{fig:multiPieFits}. We perform an unconstrained face recognition experiment over pose and expressions, see Table~\ref{tab:multiPieRecognition}. The face recognition results are competitive compared to state of the art inverse rendering techniques. Additionally we present qualitative results in a more realistic setting on the Labeled Faces in the Wild (LFW) database \cite{LFWTech} in Figure~\ref{fig:lfwFits}. For all fitting experiments we initialized the pose with 9 manually annotated landmarks.

\begin{table}[h]
\begin{tabular}{ |c|c|c|c|c| } 
\hline
probe    & $15^\circ$           & $30^\circ$           &  $45^\circ$          & smile                \\
probe id & \texttt{01\_140\_16} & \texttt{01\_130\_16} & \texttt{01\_080\_16} & \texttt{02\_050\_16} \\ 
\hline
BFM '17  & \textbf{98.8}        & \textbf{98.0}        & \textbf{90.0}        & \textbf{87.6}        \\ 
BFM '09  &  97.6                &  95.2                &  89.6                &  -                   \\ 
BU3D-FE  &  90.4                &  82.7                &  68.7                &  59.4                \\ 
\hline
\end{tabular}
\caption{Face Recognition results on the Multi-PIE database. The neutral images with 0 $^{\circ}$ of yaw angle build the gallery. We present results for the probe images over different poses and for smile. We compare the results obtained with the new BFM vs. the original BFM-2009 and the model built on the BU3D-FE dataset.}%
\label{tab:multiPieRecognition}
\end{table}

\section{Conclusion}\label{sec:conclusion}

As a first central contribution, we presented a non-rigid registration method for facial shapes based on Gaussian process registration. The framework cleanly separates domain-specific knowledge as modeled by a Gaussian process from the actual registration algorithm. We specifically demonstrated how to build a prior model for face registration by combining multiple deformation scales, symmetry and mouth opening for facial expressions using kernel modeling techniques. The pipeline has been made available open source together with the publicly available face database to reach full reproducibility based on open data. Part of the framework is a model-building pipeline, which enables the construction of a morphable model from registered data, and an inverse rendering software, which applies the built model to 2D images of faces.
Furthermore, we release a new BFM-2017 based on high-quality shape and color data with facial expressions and an improved age distribution. With qualitative and quantitative evaluations, we compared the model performance for inverse rendering and face recognition. We showed that the new Basel face model outperforms the model built on the BU3D-FE dataset and also its predecessor from 2009 \cite{paysan20093d}. With this work on face registration and the release of an open pipeline for registration, model-building and model fitting, we enable the community to reproduce and compare the results of neutral and facial expression registration, model-building and model-fitting. The pipeline code has been released on Github\footnote{https://github.com/unibas-gravis/basel-face-pipeline} and the new BFM-2017 with expressions is available for download on our website\footnote{http://gravis.dmi.unibas.ch/pmm/}.

\bibliographystyle{ieee}
\bibliography{fg-2018}

\end{document}